# Vulnerability of 3D Face Recognition Systems to Morphing Attacks

Sanjeet Vardam, Luuk Spreeuwers
*University of Twente*
s.p.vardam@student.utwente.nl, l.j.spreeuwers@utwente.nl

This work is part of the iMARS project. The project received funding from the European Union's Horizon 2020 research and innovation program under Grant Agreement No. 883356.
Disclaimer: this text reflects only the author's views, and the Commission is not liable for any use that may be made of the information contained therein.

*Abstract*—In recent years face recognition systems have been brought to the mainstream due to development in hardware and software. Consistent efforts are being made to make them better and more secure. This has also brought developments in 3D face recognition systems at a rapid pace. These 3DFR systems are expected to overcome certain vulnerabilities of 2DFR systems. One such problem that the domain of 2DFR systems face is face image morphing. A substantial amount of research is being done for generation of high quality face morphs along with detection of attacks from these morphs. Comparatively the understanding of vulnerability of 3DFR systems against 3D face morphs is less. But at the same time an expectation is set from 3DFR systems to be more robust against such attacks. This paper attempts to research and gain more information on this matter. The paper describes a couple of methods that can be used to generate 3D face morphs. The face morphs that are generated using this method are then compared to the contributing faces to obtain similarity scores. The highest MMPMR is obtained around 40% with RMMR of 41.76% when 3DFRS are attacked with look-a-like morphs.

## I. INTRODUCTION

In recent years, use of face recognition for biometric identification has become popular. Due to development in software technologies along with hardware, access to face recognition systems have become easy. This development has also allowed easier access to technologies that can exploit the shortcomings of an automated face recognition system. Face morphing is one such threat to a face recognition systems which has proven to be quite effective. The basic idea is to synthetically generate a face image from a combination of faces of two subjects. This generated image is expected to be similar to both the subjects in terms of various relevant features. It has been observed that the state-of-the-art face recognition systems can have higher false matches through introduction of such face morphs that are generated with relevant modifications [1].

This vulnerability of a face recognition system to face morphs has given rise to a whole domain of Morphing Attack Detection (MAD). It is necessary to mitigate such anomalies by understanding the effects of different morph generation techniques on performance of face recognition systems. Pertaining to MAD a considerable amount of work is done to understand these effects on a 2D face recognition system. Substantial amount of research has been done to differentiate potential parts in the training process and working mechanism of a 2D face recognition system that can be improved to mitigate Face Morphing Attacks [2], [3]. Some challenges that remain a big barrier for this problem are availability of representative, large-scale datasets and reference evaluation procedures of novel detection algorithms, addressing which will require considerable time, effort and resources [4].

Now developments in the 3D domain have become the next step for face recognition systems. With improvements in hardware to obtain 3D facial scans, it has become possible to build large 3D facial datasets. These datasets have made it possible to train and develop reliant 3D Face Recognition systems. These systems vary in approach that they follow to obtain accuracies comparable to a 2D face recognition system. With further improvement in technology it is only expected that 3D face recognition systems will be brought to mainstream and follow a development path similar to its 2D counterpart. It has the potential to eradicate some major shortcomings of 2D face recognition. An argument here is that because there is an additional information source in the form of depth or range, 3D face recognition systems can develop better feature discriminators. When comparing the vulnerability of these systems towards face morphing, an expectation can be set from 3D systems that due to the additional information, it might be more robust against face morphing. Going one step ahead, one can also ask the question, are 3D face recognition systems even vulnerable to morphing attacks? This paper makes an attempt to answer this question.

The progress in 2D face morphing has allowed substantial research for MAD in 2D face recognition systems. Comparatively, generation of 3D face morphs is a relatively untouched topic. Further using these morphs to measure their effectiveness against a 3D face recognition system is unknown. The novelty of this paper lies in the attempt to attack 3D face recognition with 3D face morphs. This paper describes a couple of ways to obtain a 3D face morph. The prime method of obtaining morphs in this paper is using an already existing pipeline developed for Large Scale 3D Morphable Models [5]. This approach allows generation of 3D face morphs in a fully automated manner, suitable for a large-scale evaluation of morphing attacks with minimal requirements. These attacks are tested specifically on face recognition systems described in papers [6] and [7].

Generation and testing of face morphs is done using FRGC [8] and Bosphorous [9] 3D facial datasets. This datasets allow the research to be done on a large number of samples obtained from many different subjects under controlled and uncontrolled environmental conditions. They allow a considerable amount of variations, to enable a selection criteria for generation of morphs. The intention behind this selection is to improve the performance of morphing attack. This is done to understand external factors that could potentially influence 3D face morphing. The selection criteria is based on scores obtained from comparing faces, with selection of faces that have high score and hence can be considered similar. This criteria allows a consistent approach on selection of subjects and gives an idea if similar looking faces generate better morphs in 3D, similar to 2D.

More relevant work about the morphing in general and different 3D morphable models is described in section II. This is followed by highlighting the objectives of this research in the form of questions. Section IV describes the methods used for generation of 3D face morphs. This is followed by a section about selection criteria that will be used to choose certain subjects in order to see their effect on the final results. The evaluation metrics section explains the metrics used to evaluate these results. And then the experiment section gives a brief overview about the experiments performed in order to answer the research questions posed in this paper. Experimental results obtained so far are mentioned here followed by a brief discussion explaining this. And then a short conclusion about the research done in this paper.

## II. RELATED WORK

This section covers an overview of work that has been done on 3D Face Morphing and work done to build generic 3D face models, wherein the approach of converting multiple facial meshes into a main model is described. Also some surveys related to 2D face morphing, give an idea of how a 3D face morphing system can be evaluated and what are the factors that need to be thoroughly checked for the evaluation. As there is less work done on 3D Face Morphing specifically, this section mainly explores the work that is relevant to the morphing process.

3D face morphing can be considered as the next step to current 2D face morphing. Hence the basis for a 3D face morphing would be relatively similar to 2D face morphing. Survey about Face Morphing Attacks presented in [10] covers different 2D morph generation techniques. It also compares landmark based morph generation to newer Deep learning based morph generations. Another paper [11] also provides an extensive overview about fundamentals of 2D face morphing. It also describes various factors to assess the quality of Face Morphs, while also covering metrics that evaluate the morphing attacks.

Early works of 3D Face Morphing began in the form of Metamorphosis on 3D head models in order to achieve new 2D images from existing 2D images [12]. The idea was to map the 3D models onto a 2D space and use these maps to obtain a morphed map from basic 2D morphing technique. This process was used to obtain a metamorphosed 3D model. A survey of 3D metamorphosis from [13] distributes the approaches for metamorphosis into volume based approach and boundary based approach. The work of Chen and Banks was applied by Steyvers to extend his 2D morphing algorithm to obtain morphs of two 3D head models [14]. The work was based on manual segmentation of facial features to create and obtain a correspondence between both 2D maps, hence obtaining a proper 3D morphed face. No further analysis was done on these results.

This was then immediately followed by Lee and Thalmann with their paper [15]. The idea was to prepare a 3D generic model, and use this to fit the feature lines obtained from the front and side view of the subject in order to obtain an individualized head along with a properly fitted texture. Hence using 2D morphing on textures along with 3D interpolation in shape, resulted in a 3D morph between two facial models.

Work done in [16] to create a generic face model i.e. a 3D Morphable model named Basel Face Model, is considered groundbreaking. This paper had a major impact on the approach of 3D facial reconstruction and synthesis. The framework for this model was presented in [17]. It performs face registration to obtain correspondence among individual meshes and then uses this with Gaussian process model as a prior to generating a morphable face model. This model is an approach to obtain linear combinations of faces in a meaningful way. Hence it becomes a powerful and convenient tool to generate face morphs through proper input.

The Large Scale Face Model presented in [5] contributes towards making the process of generating a 3D morphable model relatively easy. The idea is to have an automated process to build a 3DMM from a large collection of 3D scans. This construction is proposed in a way that requirements of input are low without any trade off between automation and model quality.

An overview about different 3D Morphable Face Models is well presented in [18]. This paper provides a detailed view of the history of 3DMMs and their progress through the past few decades. It takes an in-depth view of different processes that are involved in building 3DMMs and provides a comparison between different options at each step. This process distinctly includes Face Capturing, Modelling, Image Formation, Optimization, Applications, etc.

## III. RESEARCH QUESTION

The purpose of this research is to gain more understanding of the relation between 3D face morphing and 3D Face Recognition systems by answering following questions:

- How can a 3D face morph be obtained from two face meshes? What are different suitable approaches for this?
- Are 3D Face Recognition systems vulnerable to attacks through these face morphs?
- If so, then to what extent are 3DFR systems vulnerable?
- What factors in generation of morph influence the vulnerability of the 3DFR system?

## IV. 3D FACE MORPH GENERATION

*A. Introduction*

As discussed in II, there is not a significant amount of work done to specifically generate 3D face morphs.

In 2D face morphing techniques a common algorithm is to detect facial feature points of 2 subjects. Connecting these points yields triangular meshes. Through triangle-to-triangle correspondence the same sections in two different faces are warped to obtain intermediate shapes. For each pixel in this new shape, corresponding pixels of both face images are cross-dissolved by interpolation, as shown in Fig. 1. Many other algorithms are developed for same purpose and are broadly classified into mesh warping, field morphing, radial basis functions, thin plate splines, energy minimization and multilevel deformations [19]. All these share the following steps: feature specification, warp generation and transition control. Feature specification is the part of the system that segregates sections on face in a meaningful way. For example, using face landmarks to obtain triangular meshes as explained above. Warp generation is the process of geometrically transforming these sections and obtaining a sort of correspondence between both inputs. Once this is obtained, transition control is used as a factor to fix the rate of warping and color blending between these inputs.

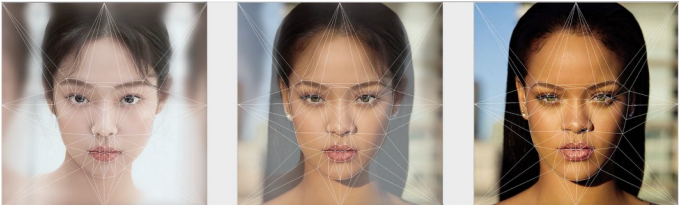

Fig. 1: 2D Face Morphing example. Original faces on left and right, with morph in the centre. Lines represent the feature specification.

The general steps involved in 2D face morphing can be modified in a relevant way to apply to 3D data as shown in Fig. 2. For 3D face morphing, the operations will stay similar to those as explained above from a broader perspective. So in order to achieve a 3D face morph, the idea is to first have a suitable method of feature specification. Once this specification is obtained in a reliable way, perform operations on them to obtain a general transformation. This transformation should be done in such a way that there is a sense of linearity for translations. This will allow for transition control to be the deciding factor for weightage of inputs in the face morph. As 3D face recognition systems used in this paper are based solely on shape data, color blending of texture is not the topic of discussion in this paper.

Two methods that are discussed in this paper that cover above steps in a proper manner. The simple way to do morphing is to just average the faces in terms of depth, so that they develop a linear mixture of two faces. Another method is to use the pipeline for building 3D Morphable Facial Models. These methods are briefly discussed ahead.

B. Background

*1) Facial 3D Morphable Model:* The Facial 3D Morphable Model(3DMM) is defined as a generative model for face shape and appearance [18]. This model is built using multiple faces to obtain a generic model that can be altered to obtain varying face shapes. Building such a model requires a dense point-to-point correspondence amongst all the facial meshes included. This dense correspondence is obtained through a registration process that is used on all the samples, and also maintained in any further steps involved. This is one way of feature specification in 3D facial morphing. Once such a correspondence is obtained, a linear combination of faces can be defined in a meaningful way. Thus allowing warp generation with transition control.

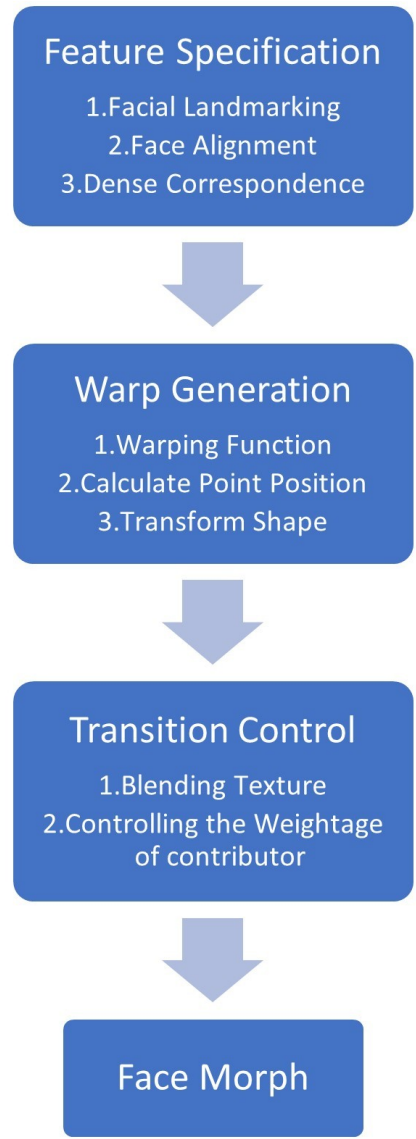

Fig. 2: Step wise distribution of 3D Face Morphing process. Feature Specification, Warp Generation, Transition Control are general steps, relevant also in 2D Face Morphing. Each step has respective sub steps, specific to 3D Face Morphing.

There are various 3DMMs built from facial databases and the pipelines for this are provided. Basel Face Model is used in this paper to test suitability of such models to obtain morphs. Then the Large Scale Facial Model pipeline is used to obtain morphs from raw datasets. The requirements for both are different due to differences in the registration process.

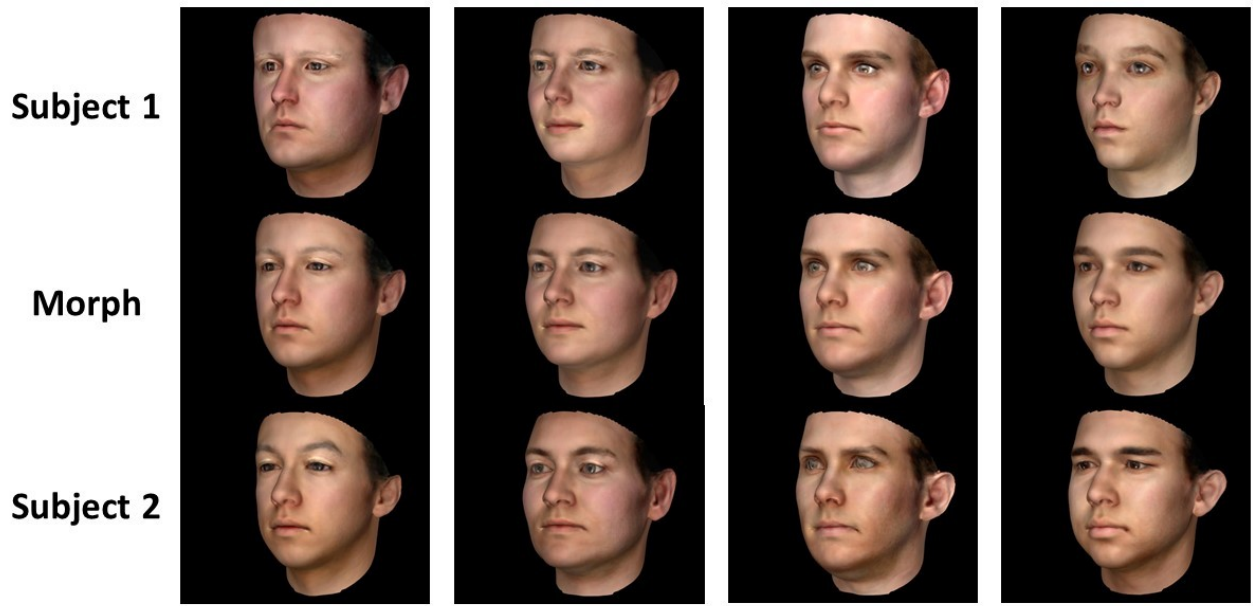

Fig. 3: Examples of averaging PCA features of two pre-registered face samples of different subjecrs to obtain a morphed face

*2) 3D Face Recognition Systems:* In order to understand the vulnerability of 3D face recognition systems against face morphing, two different 3DFRS. This is done to understand the differences in robustness of different face recognition systems against face morphing. The two systems used are based on the papers [6] and [7].

[6] is based on feature extraction using PCA-LDA by different face region classifiers. The matching score is extracted form likelihood ratio. This is a fast and accurate 3D Face Recognition system, with scores ranging from 0 to 60. 8 is the threshold and hence the scores below it are considered as a non-match. This threshold value corresponds to FMR=0.1%.

[7] presents a 3D face recognition system that is based on local shape descriptors. Feature comparison is done using cosine distance similarity metric. Hence the scores lie between 0 and 2, with anything below 0.71565 is considered a match. This threshold value corresponds to FMR=0.1%.

### C. Basel Face Model

Basel Face Model is a Facial 3DMM, which is built using the pipeline described in [17]. The latest version of this model i.e. BFM-2017 is available as an open-source. BU3D-FE along with Basel Scans are used to construct this model. This model is the base of facial 3DMMs and a lot of other pipelines are built around this specific model. Along with the model, the pipeline itself is a big contribution towards development of facial 3DMMs. The availability of this pipeline allows us to reach full reproducibility of the experiments conducted for this research.

The BFM pipeline requires 2D or 3D landmarks for registration of raw data. The registration process is carried out using a model-fitting approach. The reference surface i.e. the base model is deformed to fit the known landmarks. The landmark evaluation is done for this registration. Average distance errors range around $4 \pm 3mm$ for nose and highest being $12 \pm 6mm$ for chin.

The data provided on their website also gives access to different pre-registered face samples. They are available as coefficients, and can be used to regenerate a face. This face being obtained by fitting the BFM to 2D face images from certain datasets explained in [20]. These samples are used for initially testing the suitability of the idea of morphable models for generation of 3D face morphs as shown in Fig. 3.

### D. Large Scale Facial Model

The Large Scale Facial Model proposed in [5] was built in order to obtain a facial 3DMM with a high number of distinct facial identities, specifically 9663. The objective of this model was to attain high variance in statistical information. The major reason that LSFM is able to include such a high number of distinct samples is due to inclusion of an automated landmarking method in the pipeline used to build it. This automated landmarking of 3D face data helps overcome a huge barrier of obtaining 3D landmarks manually. Hence standard large 3D face datasets can be used to develop facial 3DMMs.

The automated landmarking uses the 2D landmarking using RGB information of the input facial mesh while preserving the 3D shape information. This helps in using the advantage of high accuracies with 2D landmarking for 3D data. Basically

RGB image from different angles is recorded for the face mesh along with XYZ shape image. The knowledge of pose and the face image helps extract landmarks using the state-of-the-art landmark localization technique. This technique is the HOG active appearance model. Thus- 68 sparse annotations are obtained automatically and used for further registration which is done using non-rigid iterative closest point (NICP) algorithm. The paper [5] also compares different registration methods to choose NICP after obtaining best results, with a mean per-vertex reconstruction error of around 1.5mm.

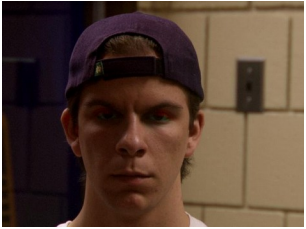

Fig. 4: Input Face images, this is texture data of the mesh which is also an input to the pipeline

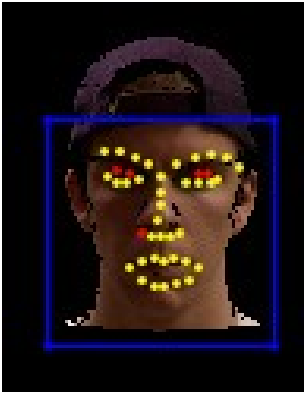

Fig. 5: Obtained facial landmarks that correspond to the 3D face. The yellow points are high confidence landmarks and red points are low confidence.

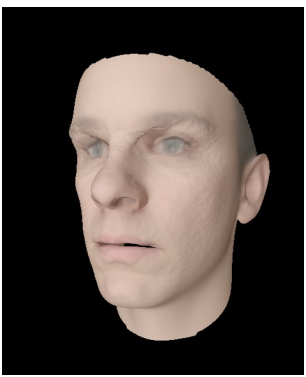

Fig. 6: Registered face model. This along with other face mesh inputs have dense correspondence, and are further used to obtain facial 3DMM

### E. Depth Averaging

This method to obtain morphs is based on the simple idea that two registered faces can be averaged in terms of their depth. This simplicity covers all three aspects of feature specification through registration. Warp generation through averaging the depth, with transition control obtained through weightage for depth values.

Registration for this method is done using the 3D Face Registration Method described in [6]. This method extracts a region of interest from the face. Then determine the vertical symmetry plane for this ROI through the nose along with finding the nose tip and slope of the nose bridge. Then transform the points in ROI to a coordinate system defined by the symmetry plane, nose tip and nose bridge. This step of transforming the point cloud into a 'nose based' coordinate system allows all the faces to correspond with at least one feature being the consistent match.

The overlapping faces now only depend on the structure of the subject face to generate an acceptable morph. The simplicity of this method has a trade off for this approach. With a necessary requirement being that both subjects need to have similar ratios for lengths on their feature points to obtain a reliable morph.

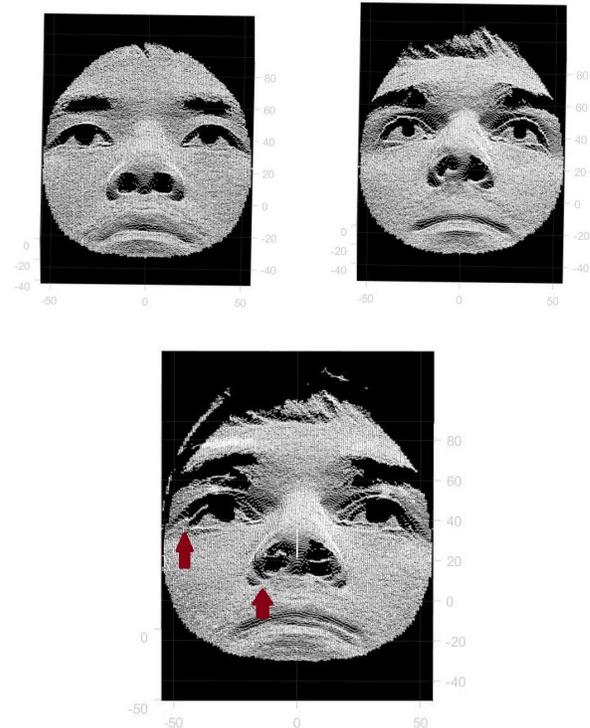

Fig. 7: Registered face mesh of two distinct subjects(above) and morphed face with depth averaging(below). The arrows in the morphed face image point to improper overlapping due to different position and size of nose, eye.

## V. EVALUATION METRICS

In order to assess the methods used for face morphing, standardized metrics are necessary. These metrics provide an understanding of the quality and effectiveness of morphs. In order to understand the effectiveness of the morph generation system, the metric should represent the ability of the system to obtain matches between generated morphs and faces of input subjects. Mated Morph Presentation Match Rate(MMPMR) and Relative Morph Match Rate(RMMR) in [21] properly quantify the vulnerability of face recognition systems to Morphing Attacks. MMPMR measures the vulnerability of system, i.e. for a certain threshold, how many % of total morphing attacks were successful for a certain set threshold. Hence this metric highly depends on the threshold. RMMR also considers FNMR in it's calculation and hence also measures the effectiveness of system against attacks along with it's vulnerability.

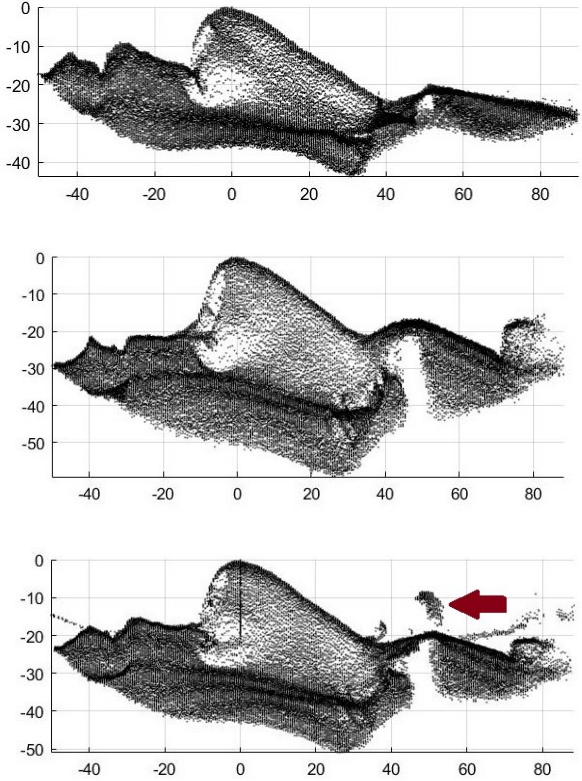

Fig. 8: From top to bottom, side view of face 1, face 2 and face morph respectively. Arrow in face morph shows some remnants formed due to averaging mesh points with a hole from the eye. Also it can be observed that all nose tips are at (0,0,0) with a morphed face showing average values for the axis on left.

## VI. EXPERIMENTS

This section explains the experiments performed for evaluation of face morphing methods described in Section IV. First experiment is generating morphs using coefficient averaging on Basel Face Model and testing them against a 3D face recognition system. Second experiment is testing the morphs generated using the LSFM pipeline, with inputs captured in an uncontrolled illumination setup. Third experiment is testing the morphs generated using the LSFM pipeline, but with inputs captured in a controlled illumination setup. Fourth experiment is testing morphs generated using depth averaging. Fifth experiment is to use selection criteria for choosing a similar set of faces to generate lookalike morphs and testing them. The face samples used for these experiments were chosen from FRGC v2.0 and Bosphorous database.

### A. Database

The experiments were performed by using either of the following data:
- Coefficients from BFM fitting algorithms
- FRGC database
- Bosphorous database

*1) Coefficients from BFM fitting algorithms:* In the initial experiment, coefficients are averaged to obtain morphs of the different faces. These coefficients are provided along with the Basel Face Model. They are obtained using the state of the art fitting algorithm described in [20]. Fitting is done on images from FERET and CMU-PIE database. This are 2D face databases. Results of fitting yields coefficients which are already provided as explained in IV-C. The coefficients are used to obtain faces of distinct subjects from the BFM. Each individual coefficient represents a certain aspect of the face, for example, width, thickness, etc.

*2) FRGC database:* FRGC v2 database is a widely used dataset for research related to 3D face recognition. It has 4007 textured 3D face scans of 466 subjects with varying facial expressions, captured under controlled as well as uncontrolled illumination conditions. The acquisition of 3D range data was done using the Minolta Vivid 900/910 series sensor.

*3) Bosphorous Database:* Bosphorous database is also a widely used dataset due to its well defined samples in terms of facial expression. It consists of 4652 textured face scans of 105 subjects and is captured in controlled illumination conditions. Each subject includes scans with 34 combinations of different expressions, poses and occlusion conditions. It is also well detailed in terms of demographics. Also each scan includes manually labelled 24 facial landmark points. The scans are captured using Inspeck Mega Capturor II 3D.

### B. Experiment 1: PCA Averaging

For this experiment, coefficients from VI-A1 were chosen for different face ids but the same lighting condition. These coefficients can be used to regenerate distinct faces. Averaging these coefficients yields a face that is expected to be the face morph. 50 faces were chosen and morphs were obtained with combinations of 2. For same face ids another set of coefficients was obtained that was fitted from a different lighting conditions. So the faces are expected to be the same but captured under different conditions. This is done with consideration of the natural biometric variance. In total 59 different subjects with 22 samples for each subject are used to compare with 2475 generated face morphs. 2475 morphs are obtained from using all pairs of face combination from 59 faces. The whole set of faces and face morphs are compared using both face recognition systems mentioned in IV-B2 for further evaluation. The bonafied 3D faces in this experiment are generated i.e. fitted. Hence this experiment only displays the possibility of 3DFRS being vulnerable to face morphs.

For simplicity, 3DFRS from [6] will be referred as likelihood ratio classifier and 3DFRS from [7] will be referred as distance similarity classifier. This are based on the scores that are obtained from them.

The results obtained from this experiment are as noted below:

*1) Distance Similarity Classifier:*

- MMPMR = 16.2%
- RMMR = 26.39%
- FNMR = 24.76%
- FMR = 1.94%

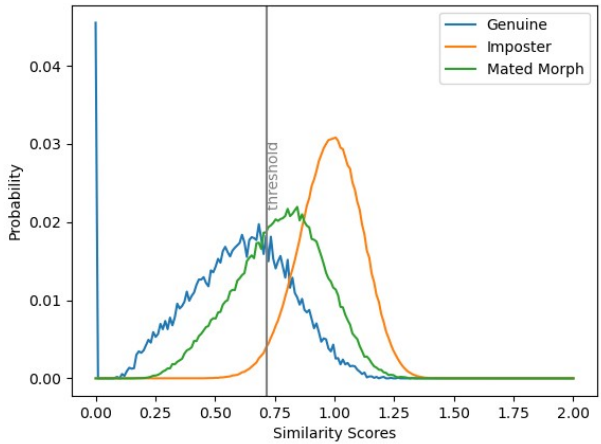

Fig. 9: Experiment 1 score distribution for Distance Similarity Classifier

*2) Likelihood Ratio Classifier:*

- MMPMR = 98.23
- RMMR = 123.14
- FNMR = 75.09
- FMR = 12.34

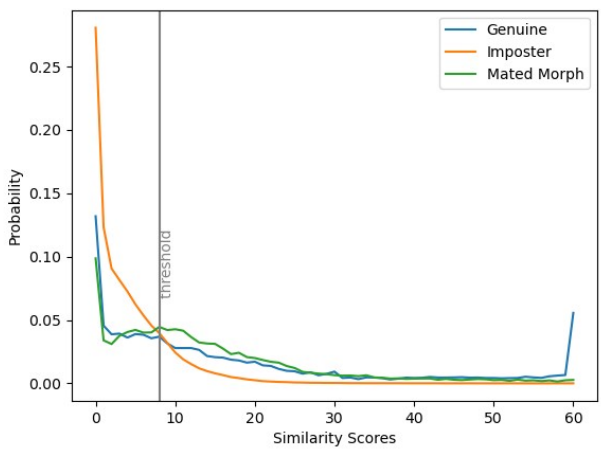

Fig. 10: Experiment 1 score distribution for Likelihood Ratio Classifier

From both the figures 9 and 10, it can be observed that high number of genuine samples do not match with each other, also confirmed by high FNMR. This is due to comparison of samples generated from fitting on 2D images of different angle, which leaves a room for error. At the same time, it can be seen that considerable amount of mated morph curve is in the matched region, which is also verified by high percentage of RMMR and MMPMR values in both cases. Due to high FMR these numbers are rendered irrelevant in terms of understanding actual vulnerability of 3DFRS, but they indeed confirm that samples of two subjects that do not match on 3DFRS, can still match with a morph that is generated from samples of both this subjects.

### C. Experiment 2: Morphs generation from uncontrolled Lighting conditions

In this experiment, morphs are generated using the LSFM pipeline. 3D face scans from FRGC databases are used as the inputs to the pipeline. The scans chosen are captured in uncontrolled illumination conditions. In total 282 combinations of subjects chosen at random from 466 subjects where used to generate morphs. This experiment is done to test if the 3DFRS are actually vulnerable to morphs generated from actual facial scans when compared against original face scans unlike the previous experiment. The resultant morphs are compared with both the face recognition system to obtain scores for further evaluation.

The results obtained from this experiment are as noted below:

*1) Distance Similarity Classifier:*

- MMPMR = 13.48%
- RMMR = 16.07%
- FNMR = 2.59%
- FMR = 0.11%

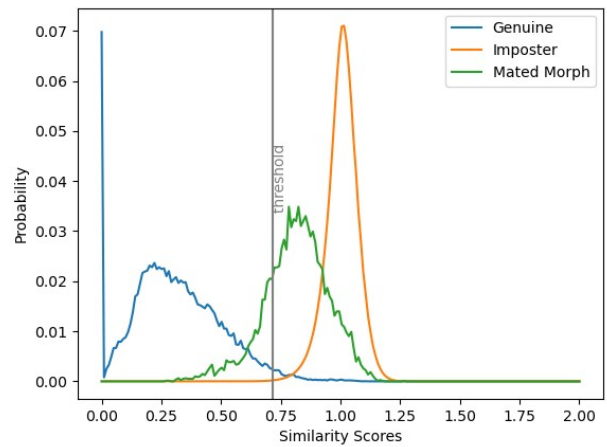

Fig. 11: Experiment 2 score distribution for Distance Similarity Classifier

*2) Likelihood Ratio Classifier:*

- MMPMR = 1.64%

- RMMR = 3.44%
- FNMR = 1.8%
- FMR = 0.1%

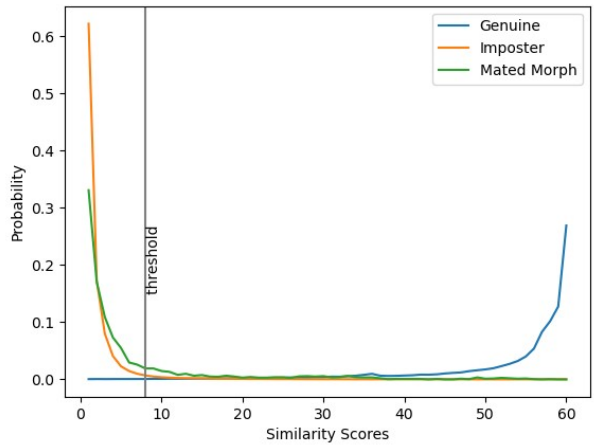

Fig. 12: Experiment 2 score distribution for Likelihood Ratio Classifier

In this experiment, it is observed that distance similarity classifier has higher MMPMR, RMMR, compared to likelihood ratio classifier. This stresses on the fact that distance similarity classifier is vulnerable to such kind of randomly generated morphs.

*D. Experiment 3: Morphs generation from controlled Lighting conditions*

In this experiment, morphs are generated using the LSFM pipeline. Data from the Bosphorous database is used in this experiment. In total 422 morphs are generated, by choosing subjects at random. For generation of morphs, neutral pose samples from individual subjects are used. Scores are obtained from comparison between this face morph and rest of the samples from the dataset. This experiment is conducted to observe, if generation of morphs from face scans obtained in controlled lighting condition, improves their quality hence resulting in matches with more samples of contributing subjects.

*1) Distance Similarity Classifier:*
- MMPMR = 0%
- RMMR = 35.25%
- FNMR = 35.25%
- FMR = 11.3%

*2) Likelihood Ratio Classifier:*
- MMPMR = 0.41%
- RMMR = 28.17%
- FNMR = 27.8%
- FMR = 36.18%

It can be seen that FMR is high in this experiment. This sheds light on the fact that a lot samples from same subjects do not match with each other. This can also be confirmed in figures 13 and 14 with genuine curve having high numbers in unmatched area. This is probably due to different expressions and poses of faces that Bosphorous dataset covers. Thus FNMR is higher for both cases which results in high RMMR, even though MMPMR is low. Which means that morphs were not effective in this case, but threshold was high for even the genuine matches.

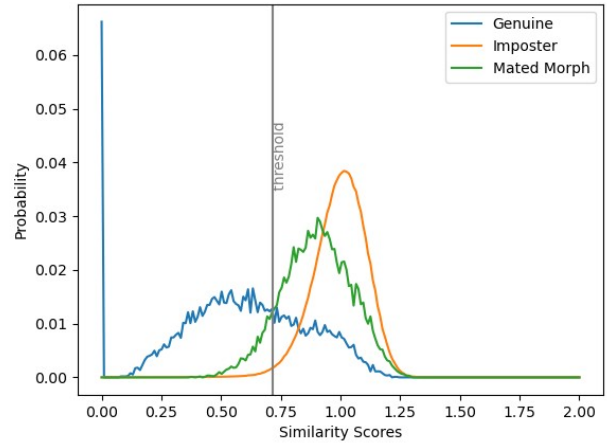

Fig. 13: Experiment 3 score distribution for Distance Similarity Classifier

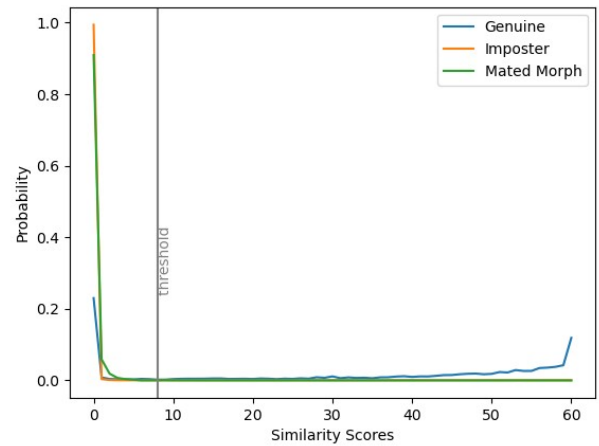

Fig. 14: Experiment 3 score distribution for Likelihood Ratio Classifier

*E. Experiment 4: Depth Averaging*

In this experiment, morphs are obtained using depth averaging methods explained in IV-E. The samples from FRGC database are chosen, irrespective of the illumination condition, as no texture is involved for this method. 560 face morphs were generated using this method from combinations of subjects chosen at random. This experiment is conducted to check vulnerability of 3DFRS against face morphs, if 3D face morph has been generated with a method as simple as depth averaging.

*1) Distance Similarity Classifier:*

- MMPMR = 2.14%
- RMMR = 4.79%
- FNMR = 2.7%
- FMR = 0.06%

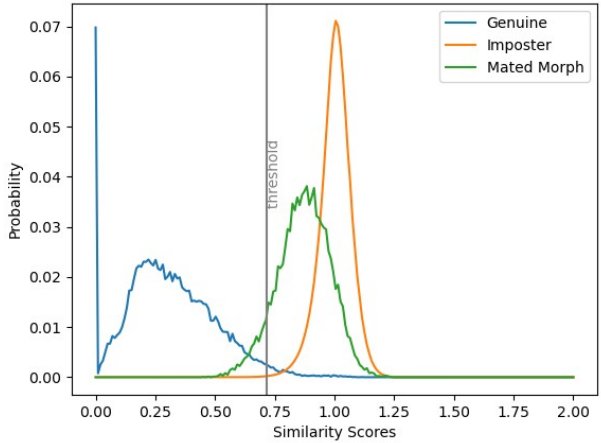

Fig. 15: Experiment 4 score distribution for Distance Similarity Classifier

*2) Likelihood Ratio Classifier:*

- MMPMR = 0%
- RMMR = 1.47%
- FNMR = 1.48%
- FMR = 0.13%

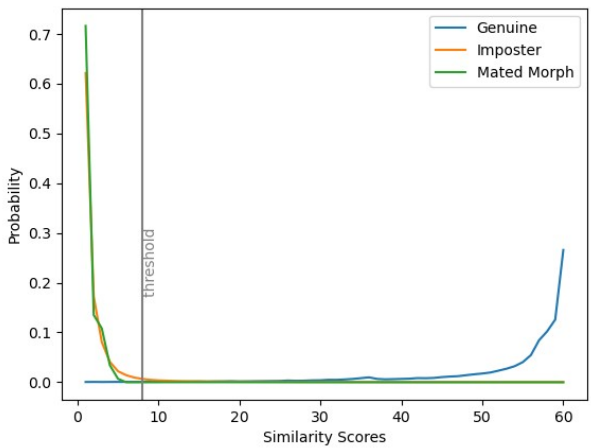

Fig. 16: Experiment 4 score distribution for Likelihood Ratio Classifier

MMPMR values obtained from experiment on both classifiers display that depth averaging is not an effective technique for morphing. This shows that some morphs obtained in this method can produce successful matches. A more manual approach with this method might yield better results on distance similarity classifier but same can't be expected to be true in likelihood ratio classifier.

### F. Experiment 5: Lookalike Morph generation

In this method a selection criteria is used to choose similar subjects for obtaining face morphs. In order to stay consistent, the selection was made based on similarity scores obtained from the 3DFR system [6]. Two subjects with scores between 3 and 7 were considered as similar. This is to keep a high threshold for faces to qualify as similar and also not consider faces that already match on the 3DFRS. In total, 872 morphs were generated from combinations obtained using this criteria. This morphs are considered look-a-like morphs and hence this would be the ideal test to check the vulnerability of 3DFRS.

*1) Distance Similarity Classifier:*

- MMPMR = 8.6%
- RMMR = 12.24%
- FNMR = 3.64%
- FMR = 0.15%

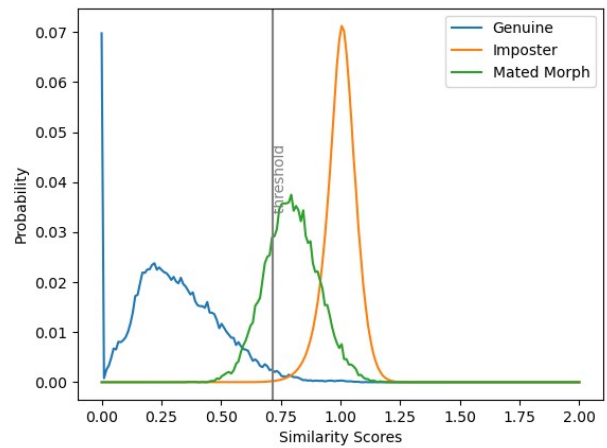

Fig. 17: Experiment 5 score distribution for Distance Similarity Classifier

*2) Likelihood Ratio Classifier:*

- MMPMR = 39.97%
- RMMR = 41.76%
- FNMR = 1.8%
- FMR = 0.28%

Based on the scores above, based on RMMR and MMPMR scores it can be observed that both classifiers are vulnerable to morphs generated from similar faces. With low FNMR, threshold for this 3DFRSs seem fine. But still relatively high number of 3D faces match with the contributing subjects.

## VII. DISCUSSION

The objective of this research was to try and find a method to obtain 3D face morphs. To find out if a 3DFR system is vulnerable to attacks from face morphs and the extent to which it can be done. Along with the understanding of factors that influence the vulnerability. In order to answer these questions, the 5 experiments explained in section VI were designed.

Experiment 1 was set up to try and figure out if 3D Morphing is possible through usage of 3D morphable models.

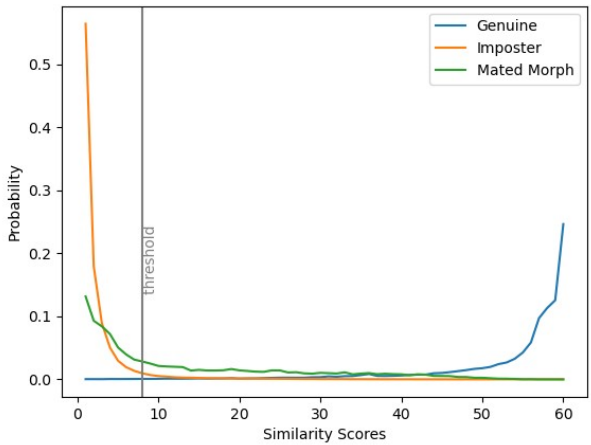

Fig. 18: Experiment 5 score distribution for Likelihood Ratio Classifier

As it can be seen in results of section VI-B, a high number of morphs in both cases seem to match with the contributing subject samples, with likelihood ratio classifier having higher amount of matches. These can be seen in high MMPMR, which means out of all the morphs, 16.2% in distance similarity classifier and 98.23% in likelihood ratio classifier, matched with at least one sample of both the contributing subjects. This just confirms the fact that 3DMM is a viable approach to face morphing. This experiment verifies nothing beyond this fact. As can be seen that FMR are above 1% in both the cases, and also the fact that all bonafied faces are generated, makes it clear that this metrics are not reliable enough to discuss further.

Experiment 2 was to use the 3D morphable model building pipeline to generate face morphs from exactly one sample of each contributing subject. These morphs show a substantially lower number of matches, which is expected. But in case of distance similarity classifier, MMPMR and RMMR are high, with low FNMR. This shows that a lot of morphs, 13.48% to be specifc, generated at random tend to match with both the contributing subjects. At the same time, even likelihood ratio classifier seems to have false matches with morphs. This experiment is based on low-quality morphs. But as the subjects were chosen at random, it can still be the case that some of the morphs were high quality. So it can be confirmed that 3DFRS in general are vulnerable to face morphing to some extent. It might be in the case that morphs are high quality.

Experiment 3 is performed to produce and test morphs, that are produced with more accuracy in terms of landmarking and correspondence. This is expected to result into more smoother morphs, with features being preserved in a better way. For this bosphorous dataset was used, which has controlled environment conditions. This may yield better automated landmarks in a easier way, with less errors. The morphs generated in this case are also from subjects chosen at random. A high FNMR depicts that many samples for same subject in this case do not match with each other. At the same time low MMPMR, almost 0, emphasizes on the fact that morphs generated in this case are not good enough. This behaviour could can be explained as followes. Both the 3DFRS from this experiments are better at distinguishing faces from Bosphorous dataset and could reduce the FNMR by using a higher threshold. This answers the research question about exploring an external factor that influences the vulnerability of 3D face morph.

Experiment 4 is performed to test if there is another suitable approach for 3D face morphing, other than using 3DMM pipeline. Depth averaging seems to be a contender for an alternative. But the experiment results shows that classifiers are highly robust against this method. It can be argued that a more manual approach with this method could yield more matches.

Experiment 5 is conducted to test extent of vulnerability of 3DFRS against high quality face morphs. The look-a-like of morphs pertains to the high similarity between both contributing faces. Results show that a high number of morphs match with both the contributing faces. This adds to the answer for the question, about what other factors influence vulnerability of 3DFRS. In this experiment, distance similarity classifier seems to be able to differentiate between such high quality morphs better than likelihood ratio classifier. But this is due to the fact that, the similar faces are chosen based on scores of likelihood ratio classifier. Hence it seems reasonable that morphs in this experiment are more likely to have successful attacks against likelihood ratio classifier. This emphasizes that 3DFRS are highly vulnerable to high-quality morphs, especially the ones that are based on the 3DFRS itself, in terms of similarity.

So the experiments performed in this paper answer all the questions asked in section III. To summarize, now we know how to obtain proper 3D face morphs. So far we know that there is only one suitable approach for it. 3DFRS are vulnerable to attacks from face morph generated using this approach. The extent of vulnerability is significant, and efforts need to be made to improve 3DFRS mentioned in this paper. One factor that influences the vulnerability of these morphs is quality of scans and other one is similarity between the faces of contributing subjects of the morph.

## VIII. Conclusion

From the experiments conducted so far, it can be seen that 3D face morphing is achievable and is able to get matches with 3DFRS. It can be seen that for attacks from look-a-like face morphs, 3DFR systems are quite vulnerable. This vulnerability is especially high when the similar faces are chosen from high scores, obtained from the 3DFRS itself, that is being attacked. In case of face scans that are generated in a highly controlled manner, the differentiation between morphs and contributing faces becomes easier. Also a method as simple as depth averaging cannot yield matches, if done in an automated way. There is still a considerable amount of work that is to be done to understand face morphing attacks against 3DFRS. Based on the results above it can be concluded that 3DFRS are vulnerable to 3D face morphing.